\documentclass{article} %
\usepackage{iclr2027_conference,times}

\usepackage{amsmath,amsfonts,bm}

\def\eqref#1{equation~\ref{#1}}

\def\1{\bm{1}}

\DeclareMathAlphabet{\mathsfit}{\encodingdefault}{\sfdefault}{m}{sl}
\SetMathAlphabet{\mathsfit}{bold}{\encodingdefault}{\sfdefault}{bx}{n}

\newcommand{\E}{\mathbb{E}}

\newcommand{\Var}{\mathrm{Var}}

\usepackage{hyperref}
\usepackage{url}
\usepackage{booktabs}       %
\usepackage{amsfonts,amsmath,amssymb,amsthm,mathtools,bm}
\usepackage{nicefrac}       %
\usepackage{microtype}      %
\usepackage{xcolor}         %
\usepackage{enumitem}      %
\usepackage{graphicx}
\usepackage{makecell}
\usepackage{colortbl}
\usepackage{algorithm}
\usepackage{algpseudocode}
\usepackage{subcaption}     %
\usepackage[capitalize,noabbrev]{cleveref}
\usepackage{wrapfig}
\usepackage{multirow}
\usepackage{titletoc}

\hypersetup{
    colorlinks,
    linkcolor={red!100!black},
    citecolor={blue!100!black},
    urlcolor={blue!80!black}
}
\usepackage[most]{tcolorbox}

\definecolor{orangesignal}{HTML}{FE6100}
\definecolor{highlightsignal}{HTML}{FFB000}
\definecolor{purplesignal}{HTML}{785ef0}
\definecolor{bluesignal}{HTML}{648fff}
\definecolor{pinksignal}{HTML}{dc267f}
\definecolor{borealisblue}{HTML}{0051A5}

\newcommand{\wh}{\widehat}
\newcommand{\defeq}{\coloneqq}

\algrenewcommand{\alglinenumber}[1]{\scriptsize #1:}

\definecolor{citationcolor}{HTML}{648fff} 
\definecolor{linkcolorcustom}{HTML}{648fff}

\hypersetup{ 
    colorlinks=true,       
    linkcolor=pinksignal,
    citecolor=citationcolor, 
    urlcolor=linkcolorcustom 
}

\colorlet{dark-green}{green!25!black}
\colorlet{dark-cyan}{cyan!25!black}
\colorlet{dark-orange}{orange!70!black}

\definecolor{blue_cblind}{HTML}{1A85FF}

\definecolor{rowalt}{gray}{0.93}                   %
\definecolor{overallbg}{RGB}{218, 230, 252}        %

\title{Amortized Off-Policy Evaluation for LLMs}

\author{%
  Younwoo Choi\textsuperscript{1,2,3,*} \quad
  Leo Feng\textsuperscript{1} \quad
  Vincent Liu\textsuperscript{1} \quad
  Haanvid Lee\textsuperscript{1,\dag} \\[3pt]
  \textsuperscript{1}RBC Borealis \quad
  \textsuperscript{2}University of Toronto \quad
  \textsuperscript{3}Vector Institute \\[3pt]
  \texttt{ywchoi@cs.toronto.edu} \quad
  \texttt{leo.feng@borealisai.com} \\
  \texttt{vincent.liu@rbc.com} \quad
  \texttt{haanvid.lee@rbc.com} \\[3pt]
  {\small \textsuperscript{*}Work done during an internship at RBC Borealis. \quad
  \textsuperscript{\dag}Corresponding author.}
}

\iclrfinalcopy %
\begin{document}

\maketitle

\begin{abstract} 
Accurate evaluation is central to selecting which LLM to deploy, yet testing a candidate on live traffic exposes real users to an unvetted model. Teams therefore evaluate candidates offline, on data produced by already-deployed models. This is off-policy evaluation (OPE), and it faces two distribution shifts: as a model is updated in post-training, its responses diverge from the logged ones (policy shift), and the reward definition under which it is judged changes with business requirements (reward shift). Classical OPE methods are ill-suited to this continual-deployment setting because they are defined per task and require fitting from scratch on every new logged dataset or reward definition. To address this, we propose PFN-OPE, a prior-data fitted network that amortizes OPE across a distribution of contextual-bandit tasks. We pretrain it once on tasks constructed from a pool of LLM responses scored by several reward functions, in which both shifts occur. At test-time it maps a logged dataset and one sampled target response per prompt to a value estimate in a single forward pass, with no per-task fitting. On HelpSteer2 and UltraFeedback with Qwen, Llama, and Gemma policies, PFN-OPE achieves 2.0 to 9.3 times lower error than the best baselines across all tested configurations in the reward-shifted settings.
\end{abstract}

\section{Introduction}
\label{sec:intro}

Choosing which large language model (LLM) to deploy hinges on accurate evaluation \citep{zheng2023judging}, yet testing a candidate on live traffic exposes users to an unvetted model. As such, candidates are often evaluated offline with logged data from previously deployed LLMs. A company has human-rated chat logs and wants to estimate the reward a candidate model would receive on the same conversations. The logs measure the deployed assistant, not the candidate. The candidate's responses differ from the logged ones, which we call \emph{policy shift}, and the rubric the human annotators apply changes with the company's requirements, which we call \emph{reward shift}. Answering the company's question from these logs is off-policy evaluation.

Off-policy evaluation (OPE) for LLM evaluation can be viewed as OPE on a contextual bandit, a one-step interaction in which a context arrives, a policy samples an action, and a reward is observed \citep{dudik2011doubly,wang2017optimal}. For an LLM, the context is the prompt, including any conversation history, the action is the response, and the reward is assigned to the response by an annotator or a reward model \citep{ouyang2022training}. The deployed model that produced the logs is the behavior policy, the candidate is the target policy, and its value is the expected reward it earns on the prompts. OPE estimates that value from the logs, which hold a reward for every behavior response and none for the response the target policy would give to the same prompt.

Three classical estimators exist — inverse propensity scoring (IPS), doubly robust (DR), and the direct method (DM) \citep{jiang2016doubly,strehl2010learning,dud}— and the first two are out of reach for many popular LLMs. IPS and DR estimators need the probability of each logged response under both policies, which a black-box model does not expose, and the target policy's responses almost never recur in the logs, so the weights would have little to no support \citep{saito2022off}. LLM evaluation therefore uses the DM: train a reward model on the logs and apply it to the target policy's responses \citep{ouyang2022training,slfrmo}. This has two costs. First, every new logged dataset or reward definition means training a new reward model. Each checkpoint of a post-training run is judged under a rubric that shifts with requirements, so each is a fit from scratch. Second, no reward model can be checked on the target policy's responses, whose rewards are unobserved — exactly where it must extrapolate. Both costs come from fitting one model to one logged dataset at a time.

Prior-data fitted networks (PFNs) remove the per-dataset fit \citep{muller2022transformers,hollmann2023tabpfn,tabpfnv2}. A PFN is a transformer pretrained on datasets sampled from a prior over tasks; given a new dataset and a query in context, it returns a prediction in one forward pass, so a new dataset is a new input rather than new training data. For OPE this offers two things the direct method lacks. The first is amortization \citep{cranmer2020frontier}: each new logged dataset is a new input to the same network. The second is generalization across the shifts: a reward model fitted to the logs must extrapolate to responses it has never seen, whereas a network pretrained on many tasks in which the target policy differs from the behavior policy has seen such shifts and learned how far the target value departs from the logged rewards. Neither comes for free, because a PFN is only as good as its prior, and no prior for OPE exists. The question is whether one can build a prior whose tasks capture the reward and policy shifts LLM evaluation faces.

We propose \emph{prior-data fitted network for off-policy evaluation \textbf{PFN-OPE}} that maps a logged dataset and one sampled response per prompt from the target policy to an estimate of the target policy's value, in one forward pass and with no per-task fitting. Our key insight is that the label a PFN needs, the target policy's value, is unobservable on real logs but computable on tasks built from a pool of stored responses with rewards: when every policy is a distribution over the pool, every value is a weighted sum of stored rewards. We build the pool once, from several LLMs and several reward functions, and sample tasks from it in which both shifts occur: affine combinations of the reward functions play the role of new reward definitions, and reward-tilted reweightings of an LLM's responses play the role of post-trained policies. Pretrained on these tasks, the network learns to read the shift off the observation itself, how the target responses differ from the logged ones and what the logged rewards say under the current definition, and by the PFN argument it approximates the minimum-MSE estimator for tasks drawn from the prior.

Our contributions are:
\begin{itemize}[leftmargin=1.5em]
    \item \textbf{Amortized OPE via prior-data fitted in-context learning.} One network maps a logged dataset and sampled target responses to a value estimate in one forward pass, with no per-task fitting. This is a new problem setting for PFNs: the input is a set of paired responses, one with a reward and one without, and the label is a functional of the whole set rather than the label of one query (\cref{sec:estimator}).
    \item \textbf{A task prior for LLM off-policy evaluation.} Tasks are built from a pool of LLM responses with rewards from several reward functions: affine combinations of the reward functions as reward definitions, reward-tilted and routed policies as policy shifts, and target values computed from the pool (\cref{sec:prior}).
    \item \textbf{Evaluation with LLM policies.} On HelpSteer2 and UltraFeedback, PFN-OPE has the lowest MSE under all five reward definitions, four of them absent from pretraining, by $2.0$ to $9.3$ times over the best of the baselines. It has the lowest MSE in 11 of 12 behavior--target pairs when the target LLM and the reward model are both absent from pretraining (\cref{sec:experiments}).
\end{itemize}

\section{Background}
\label{sec:problem}

\paragraph{Off-policy evaluation.}
We consider off-policy evaluation (OPE) in a contextual-bandit setting. In a
contextual-bandit interaction, the environment provides a context
$x\in\mathcal X$, a policy samples an action $a\in\mathcal A$, and the
environment returns a reward $r$ with a reward function $R:\mathcal X \times \mathcal A \rightarrow \mathbb{R}$. The behavior
policy $\mu:\mathcal X\rightarrow \Delta(\mathcal A)$ samples the logged data $\mathcal D_\mu
\defeq
\{(x_i,a_i,r_i)\}_{i=1}^{n}$ from an environment, while the target policy $\pi$ is the
policy whose value we want to estimate. OPE estimates the target policy value $V(\pi)$ as the
expected reward of $\pi$: $V(\pi)
\defeq
\mathbb E_{x\sim p(\cdot), a\sim\pi(\cdot\mid x)}
\left[R(x,a)\right]$.

The main challenge in OPE is that the behavior and target policies may induce
different action distributions, resulting in distribution shift
\citep{saito2022off}. The logged data contain
rewards only for actions sampled by the behavior policy, leaving the rewards
of actions sampled from the target policy unobserved. The distribution shift becomes more severe as the behavior and target
policies diverge.

In our application, $x$ is a prompt, $a$ is a generated response, and $r$ is the reward assigned to that
response by a human annotator. The behavior policy $\mu$ is the LLM that produced the logged responses, and $\pi$ is the LLM being evaluated.

\paragraph{Off-policy evaluation estimators.}
OPE estimators fall into three methods: inverse propensity scoring (IPS), direct method (DM), and doubly robust (DR) estimation \citep{jiang2016doubly}. IPS estimates the target policy value by correcting the behavior policy induced action distribution to that of the target policy by reweighting logged rewards with IPS weights \citep{saito2022off}. IPS is unbiased given behavior policy. However, its variance grows as the action space increases. The direct method approximates the reward function from logged data and therefore has low variance but high bias due to estimation error from distribution shift. DR combines the advantages of both IPS and DM \citep{farajtabar2018more}, doubly robust estimator is unbiased when either IPS or DM is unbiased and has lower variance compared to IPS.

Conventional OPE estimators do not amortize inference across evaluation
datasets. For IPS and DR, they must recompute importance weights for each behavior--target policy pair and require policy probability information that may be unavailable for black-box LLMs. DM also requires re-training when the ground-truth reward function changes.

Our work focuses on improving the DM's ability to predict rewards
for target policy responses with amortized off-policy evaluation capability. This focus is motivated by the high variance of
IPS in the large response spaces induced by LLM policies. The resulting reward model can also serve as the DM component of a DR estimator when combined with an IPS method.

\paragraph{Prior-data fitted networks.}
A prior-data fitted network (PFN) is a single transformer $q_\theta$ pretrained across many datasets: given a dataset and a query in context, it outputs a predictive distribution for the query's label in one forward pass, so a new dataset is a new input rather than new training data \citep{muller2022transformers}. The datasets come from a synthetic prior over tasks $p(\tau)$. In the standard PFN a task $\tau$ is a relationship between inputs $z$ and labels $y$, so sampling $\tau\sim p(\tau)$ samples a function; from $\tau$ we draw a dataset $\mathcal D\defeq\{(z_i,y_i)\}_{i=1}^{n}$ and a query $z_*$ with hidden label $y_*$. Pretraining minimizes the negative log-likelihood of the hidden label,
\begin{equation}
\mathcal L_{\mathrm{PFN}}(\theta)\defeq\E_{\tau\sim p(\tau)}\,\E_{(\mathcal D,z_*,y_*)\sim p(\cdot\mid\tau)}\bigl[-\log q_\theta(y_*\mid\mathcal D,z_*)\bigr],
\label{eq:pfn-loss}
\end{equation}
an expectation over tasks and, within each task, over the dataset and query drawn from it.

The property PFN-OPE inherits is that this loss is minimized by the posterior predictive, ${q^\star(y_*\mid\mathcal D,z_*)=p(y_*\mid\mathcal D,z_*)}={\int p(y_*\mid z_*,\tau)\,p(\tau\mid\mathcal D)\,d\tau}$, whose mean is the minimum-MSE estimate of $y_*$ for tasks drawn from the prior \citep{muller2022transformers}. A trained PFN thus performs Bayesian inference under $p(\tau)$ without representing the posterior, and the prior is its entire inductive bias: which tasks the prior contains decides what the network infers from a small dataset and how far it transfers. For a real-valued label, a PFN's output is a categorical distribution over fixed intervals of the label's range \citep{muller2022transformers}. PFN-OPE keeps this pretraining recipe and the Bayes argument. \cref{sec:method} changes what a task, a dataset, and a prediction target are.

\section{PFN-OPE: Prior-Data Fitted Network for Off-Policy Evaluation}
\label{sec:method}

We propose PFN-OPE for efficient off-policy evaluation of LLM policies under
policy and reward shifts. Unlike conventional OPE methods, which fit an
estimator separately for each task, PFN-OPE amortizes estimation across a
prior of OPE tasks, allowing a single pretrained network to estimate
target policy values without per-task fitting. We first formulate OPE as
in-context prediction in \Cref{sec:estimator}, then define the task prior
used to generate pretraining tasks in \Cref{sec:prior}, and finally describe
the transformer architecture in \Cref{sec:architecture}.

\subsection{OPE as in-context prediction}
\label{sec:estimator}

PFN-OPE formulates OPE as an in-context prediction problem in which the
target policy value is estimated from a behavior policy sampled logged data and target policy
sampled target responses. 

\paragraph{Input.} We assume access to a logged dataset
$\mathcal{D}_\mu\defeq\{(x_i,a_i^\mu,r_i^\mu)\}_{i=1}^n$, where $x_i$ is a
prompt, $a_i^\mu$ is the response sampled from the behavior policy
$\mu$, and $r_i^\mu=R(x_i,a_i^\mu)$ is its observed reward.
We also assume black-box sampling access to the target policy $\pi$. For each
prompt $x_i$, we sample one target response
$a_i^\pi\sim\pi(\cdot\mid x_i)$ and form the target prompt--response dataset
$\mathcal{D}_\pi\defeq\{(x_i,a_i^\pi)\}_{i=1}^n$. Together,
$\mathcal{D}_\mu$ and $\mathcal{D}_\pi$ constitute the OPE observation
$O\defeq(\mathcal{D}_\mu,\mathcal{D}_\pi)$, which serves as the input to the
PFN-OPE transformer. Equivalently, the input can be represented as  
$O = \{(x_i,a_i^\mu,r_i^\mu,a_i^\pi)\}_{i=1}^n$. The target-response rewards are not
included in $O$ and must be inferred by the model.

\paragraph{Prediction target.} PFN-OPE is trained to predict the correction value $C$ in \cref{eq:correction} given $O$ rather than predicting target-policy value directly. Subtracting the average logged reward from $V(\pi)$ centers the target on the
observed behavior reward and focuses prediction on the target-policy correction. 
\begin{equation}
C\defeq V(\pi)-\bar r,
\qquad
\bar r\defeq\frac1n\sum_{i=1}^{n}r_i .
\label{eq:correction}
\end{equation}

\paragraph{Pretraining.} 
During pretraining, PFN-OPE learns an inference rule across various OPE tasks. Tasks $\tau$ are sampled from the task prior described in \cref{sec:prior}. A sampled task $\tau$
specifies a reward function, and behavior and target policies. Given $\tau$, we generate the observation dataset $O\sim p(O\mid\tau)$ for pretraining. Because reward functions and policies are known, PFN-OPE can be supervised to predict the correction target and consequently the target policy value.

We train the network via quantile regression. A quantile head returns $\widehat Q_\theta^{\kappa}(O)$, an estimate of the $\kappa$-quantile of $C$ given $O$, for every level $\kappa$ in $\mathcal K$, and pretraining minimizes the pinball loss in \cref{eq:training-loss} \citep{koenker1978regression}. We follow TabICLv2 \citep{qu2026tabiclv2}, which found quantile regression more accurate than TabPFN's discretized output for regression targets. Another advantage is that a categorical head over fixed intervals \cref{sec:problem} fixes the range of $C$ before pretraining, whereas the quantile levels adapt to each task's predictive distribution. 

For each level $\kappa$, the expected pinball loss is minimized by the $\kappa$-quantile of the posterior predictive $p(C\mid O)$ induced by the prior \citep{koenker1978regression}, by the argument behind \cref{eq:pfn-loss} with $(O,C)$ in place of $(\mathcal D,z_*,y_*)$. A network that attains the minimum is therefore the posterior-quantile estimator for the prior, and the prior is its inductive bias, as in \cref{sec:problem}.

\begin{equation}
\mathcal L(\theta)\defeq
\E_{\tau\sim p(\tau)}\,\E_{O\sim p(O\mid\tau)}
\Bigl[\frac{1}{|\mathcal K|}\sum_{\kappa\in\mathcal K}\ell_\kappa\bigl(C-\wh Q_\theta^{\kappa}(O)\bigr)\Bigr],
\qquad
\ell_\kappa(u)\defeq u\,\bigl(\kappa-\mathbf 1\{u<0\}\bigr).
\label{eq:training-loss}
\end{equation}

\paragraph{Inference.}
At test-time, PFN-OPE receives an observation $O$ from a test-time OPE task.
The target-policy rewards and value are unknown. The pretrained network uses
$O$ to predict the correction. In a single forward pass, it produces the
quantile estimates
$\{\wh Q_\theta^\kappa(O)\}_{\kappa\in\mathcal K}$. We aggregate these
quantiles as $\wh C_\theta(O)$ and estimate the target policy value $\wh V_\theta(\pi;O)$:
\begin{equation}
\wh C_\theta(O)\defeq\frac{1}{|\mathcal K|}\sum_{\kappa\in\mathcal K}\wh Q_\theta^{\kappa}(O),
\qquad
\wh V_\theta(\pi;O)\defeq\bar r+\wh C_\theta(O).
\label{eq:value-readout}
\end{equation}

Averaging the predicted quantiles approximates the mean of the predictive
distribution of the correction. Hence,
$\wh V_\theta(\pi;O)$ approximates the posterior-mean estimate of the
target-policy value under the task prior, which is the minimum-MSE estimator
for tasks drawn from that prior.

PFN-OPE handles both policy and reward shifts through the
observation, without parameter updates. A policy shift changes the
target-response component $\mathcal D_\pi$ in $O$, because the new target
policy produces different responses for the logged prompts. A reward shift changes the rewards assigned to the logged
responses when they are rescored under a new reward function, while the target
responses remain unscored. Both
shifts are reflected in $O$ and processed by the same pretrained inference
rule.

\subsection{Constructing Task Prior from a Response Pool}
\label{sec:prior}

The task prior determines the reward and policy shifts during pretraining, and therefore serves as the estimator's inductive bias. Broader coverage of reward and policy shifts during pretraining generally improves PFN-OPE's ability to generalize to related shifts at test-time. Each pretraining task
must also provide a target-policy value $V_\tau(\pi)$ for supervision.

Generating and scoring new LLM responses for every pretraining task would be computationally expensive. We instead construct a finite response pool before
pretraining and reuse it across tasks. Each task is obtained by selecting a
prompt distribution $p(x\mid\tau)$, a reward function $R_\tau$, a
behavior policy $\mu$, and a target policy $\pi$. The target-policy value
$V_\tau(\pi)$ can then be evaluated as a finite sum over the pool, allowing us
to sample pretraining tasks without additional LLM generation or reward-model
evaluation.

\paragraph{Response pool.}
We assume that a set of LLMs $\mathcal G$ is available for pretraining. An LLM $g\in\mathcal G$ is a model's parameters together with its system prompt and decoding settings, and $g(a\mid x)$ is its response distribution. For every prompt $x$ in a finite set $\mathcal X_0\subset\mathcal X$ and every $g\in\mathcal G$, we sample $K$ responses from $g$ once and store them as $\mathcal A_g(x)\defeq\{a_g^{(1)}(x),\ldots,a_g^{(K)}(x)\}$. Reward functions $R_1,\ldots,R_J$, the heads of one or more reward models, assign a reward to every stored response, so the pool is the set of tuples $(x,g,a,R_1(x,a),\ldots,R_J(x,a))$.

\paragraph{Reward functions.}
A task's reward function is an affine combination of the stored rewards, each standardized over the pool, $\tilde R_j\defeq(R_j-m_j)/s_j$ with $m_j$ and $s_j$ the mean and standard deviation of $R_j$ on the pool:
\begin{equation}
R_\tau(x,a)\defeq\sum_{j=1}^{J}\lambda_{\tau j}\,\tilde R_j(x,a),
\qquad
\lambda_\tau\sim\mathcal N\bigl(w,\ \operatorname{diag}(w)-ww^\top\bigr),
\label{eq:reward-prior}
\end{equation}
with evaluation weights $\lambda_\tau$ drawn once per task from a Gaussian centered on base weights $w\in\Delta^{J-1}$. The covariance is singular, so every draw satisfies $\sum_j\lambda_{\tau j}=1$.
We standardize the rewards so that the weights compare criteria on one scale, and we allow negative weights because an evaluation criterion can penalize an aspect, as helpfulness minus verbosity does. The same weights apply to every prompt and response, so the logged rewards of a task describe one consistent evaluation criterion, and each new draw of $\lambda_\tau$ is a new reward definition obtained from the stored rewards alone.

\paragraph{Behavior and target policies.}
A task's policies are distributions over stored responses. For LLM $g$ and prompt $x$, the pool policy is the uniform distribution on $\mathcal A_g(x)$,
\begin{equation}
\wh g(a\mid x)\defeq\frac1K\sum_{k=1}^{K}\mathbf 1\bigl\{a=a_g^{(k)}(x)\bigr\},
\label{eq:pool-policy}
\end{equation}
the $K$-sample Monte Carlo approximation of $g(a\mid x)$. A task picks a behavior LLM $g\in\mathcal G$ and a target LLM $g'\in\mathcal G$ and sets $\mu\defeq\wh g$. The target policy starts from $\wh{g}'$ and is shifted in the way reward-guided post-training shifts a policy, toward the responses a reward model prefers \citep{ouyang2022training}. The preference is a selection reward $S_\tau(x,a)\defeq\sum_{j=1}^{J}\omega_{\tau j}\tilde R_j(x,a)$ with selection weights $\omega_\tau\in\Delta^{J-1}$, and it acts through a selection strength $\eta\ge0$ and a routing probability $\alpha\in[0,1]$:
\begin{equation}
\nu_\eta(a\mid x)\defeq\frac{\wh{g}'(a\mid x)\,e^{\eta S_\tau(x,a)}}{\sum_{a'}\wh{g}'(a'\mid x)\,e^{\eta S_\tau(x,a')}},
\qquad
\pi\defeq(1-\alpha)\,\mu+\alpha\,\nu_\eta .
\label{eq:routing-policy}
\end{equation}
The tilt $\nu_\eta$ has the form of the optimal policy of a KL-regularized reward objective \citep{rafailov2023direct}, which is how post-training moves a policy: larger $\eta$ favors responses with higher $S_\tau$, and $\eta=0$ with $\alpha=1$ returns $\wh{g}'$ itself, a change of the LLM with no reward guidance. Routing mixes $\nu_\eta$ with $\mu$ because the value gap $V_\tau(\pi)-V_\tau(\mu)$ is then linear in $\alpha$, so the prior covers shifts from none ($\alpha=0$) to the full gap ($\alpha=1$); a target response routed to $\mu$ is a fresh draw from $\wh g$, not a copy of the logged response. The selection weights $\omega_\tau$ are drawn independently of $\lambda_\tau$ because the reward that steered a policy need not be the reward under which it is judged.

\paragraph{Training labels.}
A task fixes its prompt distribution $p(x\mid\tau)$ on $\mathcal X_0$ before any observation is drawn. The value of the target policy $\pi$ at a particular task $\tau$ is computed as
\begin{equation}
V_\tau(\pi)=\sum_{x\in\mathcal X_0}p(x\mid\tau)
\sum_{a}\pi(a\mid x)\,R_\tau(x,a).
\label{eq:exact-task-value}
\end{equation}

The label $C=V_\tau(\pi)-\bar r$ of \cref{eq:correction} is defined relative to the observed logged mean, so two observations of one task carry different labels, and the network learns to predict the correction for the dataset in hand rather than the task's average gap. \Cref{alg:overview} summarizes pretraining.

\subsection{PFN-OPE Architecture}
\label{sec:architecture}

PFN-OPE follows the PFN framework, but adapts its
transformer architecture to the structure of the OPE observation $O$. In
particular, we modify three components of the standard PFN architecture:
the input representation, the attention mask, and the readout.

\paragraph{Input representation.}
The input is the OPE observation
$O$, which is represented by two tokens:
\begin{equation}
L_i=f_\mu\bigl(\phi_x(x_i),\phi_a(a_i^\mu),r_i^\mu\bigr),
\qquad
Q_i=f_\pi\bigl(\phi_x(x_i),\phi_a(a_i^\pi),\mathrm{MASK}\bigr),
\label{eq:input-tokens}
\end{equation}
where $L_i$ represents the logged response and its observed reward, whereas
$Q_i$ represents the target response whose reward is unobserved. Prompts and
responses are mapped to frozen text embeddings $\phi_x(x)$ and $\phi_a(a)$
using Qwen3-Embedding-4B \citep{qwen3embedding}. 
The symbol $\mathrm{MASK}$ denotes a learned vector that replaces the
unobserved target reward.

\paragraph{Attention mask.}
For the input structure, we use a directional attention mask. Each logged token $L_i$ attends only to the logged tokens, whereas each target token $Q_i$ attends to all logged and target tokens. Target tokens attend to the logged tokens because the reward of a target response must be inferred from the logged responses and their observed rewards. They also attend to each other because PFN-OPE predicts the correction $C$, which depends jointly on the entire set of sampled target responses.

\textbf{Readout.}
The transformer includes a learned summary token whose final representation
is passed to an output head that predicts the quantiles
$\{\widehat{Q}_\theta^\kappa(O)\}_{\kappa\in\mathcal{K}}$ of the correction
$C$. We obtain $\widehat{C}_\theta(O)$ by averaging these predicted
quantiles, as defined in Equation~\eqref{eq:value-readout}.

\begin{figure}[t]
\centering
\includegraphics[width=0.85\linewidth, trim=0 0 0 0, clip]{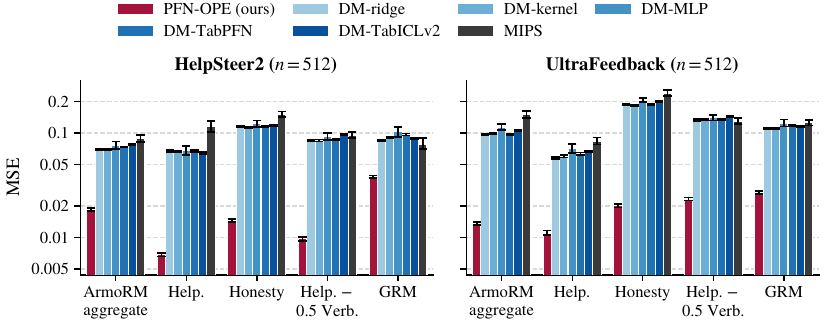}
\vspace{-4pt}
\caption{\textbf{One checkpoint transfers to four rewards absent from pretraining.} MSE at $n=512$ over the six pairs of seen LLMs, with 95\% intervals. The ArmoRM aggregate is the seen reward; the other four are unseen. PFN-OPE's MSE is $2.0$ to $9.3$ times lower than the best baseline's.}
\label{fig:reward-shift}
\end{figure}

\section{Experiments}
\label{sec:experiments}
\vspace{-4pt}
The experiments ask whether PFN-OPE replaces per-task fitting under the \emph{reward shift} and \emph{policy shift}. The metric is the MSE between $\wh V(\pi)$ and $V_\tau(\pi)$ \citep{saito2021open}. Five questions organize the section:
\begin{enumerate}[leftmargin=2.4em, itemsep=0pt, topsep=1pt]
    \item[\textcolor{bluesignal}{\textbf{Q1}}] Does PFN-OPE generalize to rewards absent from pretraining? (\cref{sec:exp-reward})
    \item[\textcolor{bluesignal}{\textbf{Q2}}] Does it generalize to unseen target LLMs under an unseen reward model? (\cref{sec:exp-joint})
    \item[\textcolor{bluesignal}{\textbf{Q3}}] Mechanism: why does PFN-OPE win, and how much logged data does it need? (\cref{sec:exp-correction-n})
    \item[\textcolor{bluesignal}{\textbf{Q4}}] Prior ablation: which axis of the prior carries the gain? (\cref{sec:exp-ablation})
    \item[\textcolor{bluesignal}{\textbf{Q5}}] Prior coverage: what do the shifted policies add? (\cref{sec:exp-coverage})
\end{enumerate}

\subsection{Common experimental setup}
\label{sec:exp-setup}
\paragraph{Pool, prompts, and LLMs.}
Prompts come from HelpSteer2 \citep{wang2024helpsteer} and UltraFeedback \citep{cui2024ultrafeedback}, reported separately. Each dataset holds out $1{,}024$ evaluation prompts that never enter pretraining. Each LLM contributes $K=8$ responses per prompt. Three seen LLMs form the pool $\mathcal G$ of \cref{sec:prior}: Qwen2.5-1.5B-Instruct, Qwen2.5-7B-Instruct \citep{qwen25technicalreport2024}, and Llama-3.2-1B-Instruct \citep{llama3herd2024}. Two unseen LLMs serve only as evaluation targets: Llama-3.2-3B-Instruct, a larger checkpoint of a seen family, and Gemma-3-1B-IT \citep{gemmateam2025gemma3technicalreport}, an unseen family. Every evaluation policy is the pool policy of one LLM (\cref{eq:pool-policy}), so policy shift in evaluation is a change of LLM.
\vspace{-7pt}
\paragraph{Rewards.}
ArmoRM \citep{wang2024armorm} outputs 18 sub-rewards and the aggregated value. The prior's reward functions $R_1,\ldots,R_J$ in \cref{eq:reward-prior} are this aggregate and 15 of ArmoRM's objectives. Four rewards are unseen. We hold out three from ArmoRM: helpfulness, honesty, and helpfulness minus $0.5$ times verbosity. The fourth is GRM-Gemma2-2B \citep{yang2024grm}, written as GRM, a different reward model. All rewards are standardized with pretraining statistics.
\vspace{-4pt}
\paragraph{Evaluation tasks and intervals.}
For each setting of dataset, a behavior LLM, a target LLM, and a reward, there are 100 tasks. A task draws $n$ evaluation prompts. At each prompt, it takes one stored behavior response with its reward and one stored target response without. This is the observation $O$, and every estimator receives the same one. The true target value is the mean reward of the target LLM's $1{,}024\times8$ stored responses. We report the MSE over the 100 tasks with a 95\% interval.
\vspace{-0.1cm}
\paragraph{Baseline estimators.}
PFN-OPE is one checkpoint, never updated for any reward, policy, or $n$. We compare it with six estimators on the same frozen embeddings. Three direct methods refit a reward regressor on every logged dataset, the per-task fitting we replace: DM-ridge, DM-kernel, and DM-MLP \citep{dudik2011doubly,wang2017optimal}. Two direct methods, DM-TabPFN \citep{hollmann2023tabpfn} and DM-TabICLv2 \citep{qu2026tabiclv2}, predict the target rewards in context with frozen tabular foundation models. They are the closest baselines: they amortize the regressor, whereas PFN-OPE amortizes the OPE. MIPS \citep{saito2022off} is marginalized importance sampling with a density ratio estimated from action embeddings.

\definecolor{pfnshade}{HTML}{DCEEEE}
\definecolor{pfnci}{HTML}{62676B}

\newcommand{\pfnv}[2]{#1\,{\scriptsize\color{pfnci}\(\pm\)#2}}
\newcommand{\pfnb}[2]{\textbf{#1}\,{\scriptsize\color{pfnci}\(\pm\)#2}}

\begin{table}[t]
  \caption{\textbf{PFN-OPE achieves the lowest MSE in 11 of 12
  behavior--target pairs when both the target LLM and reward model are
  absent from pretraining.} MSE ($\times 10^{-2}$) at $n=512$ under GRM;
  the small grey number is the larger half-width of the 95\% interval.
  Bold marks the lowest MSE within each dataset and column, determined
  before rounding.}
  \label{tab:joint-shift}
  \centering
  \footnotesize
  \setlength{\tabcolsep}{2.5pt}
  \renewcommand{\arraystretch}{1.0}
  \begin{tabular}{@{}l*{6}{c}@{}}
    \toprule
    & \multicolumn{3}{c}{\textbf{Target: Llama-3.2-3B}}
    & \multicolumn{3}{c}{\textbf{Target: Gemma-3-1B}} \\
    \cmidrule(lr){2-4}\cmidrule(lr){5-7}
    Behavior LLM
      & \shortstack{Qwen2.5\\7B} & \shortstack{Qwen2.5\\1.5B} & \shortstack{Llama-3.2\\1B}
      & \shortstack{Qwen2.5\\7B} & \shortstack{Qwen2.5\\1.5B} & \shortstack{Llama-3.2\\1B} \\
    \midrule
    \multicolumn{7}{@{}l}{\textit{HelpSteer2}} \\
    \rowcolor{pfnshade}PFN-OPE (ours)
      & \pfnb{0.041}{0.006} & \pfnb{0.87}{0.16} & \pfnb{0.97}{0.23}
      & \pfnb{1.54}{0.30}   & \pfnb{2.01}{0.21} & \pfnb{2.75}{0.42} \\
    DM-ridge
      & \pfnv{0.130}{0.047}  & \pfnv{17.78}{1.10} & \pfnv{8.45}{0.66}
      & \pfnv{5.94}{0.56}   & \pfnv{30.35}{1.90} & \pfnv{19.37}{1.80} \\
    DM-kernel
      & \pfnv{0.050}{0.055} & \pfnv{16.57}{0.91} & \pfnv{7.72}{0.87}
      & \pfnv{4.91}{0.73}   & \pfnv{30.32}{1.70} & \pfnv{18.98}{1.90} \\
    DM-MLP
      & \pfnv{1.26}{1.60}   & \pfnv{26.40}{6.60} & \pfnv{7.99}{2.00}
      & \pfnv{7.44}{3.20}   & \pfnv{39.57}{8.10} & \pfnv{20.52}{5.40} \\
    DM-TabPFN
      & \pfnv{0.055}{0.066} & \pfnv{18.17}{0.93} & \pfnv{6.99}{0.65}
      & \pfnv{4.72}{0.52}   & \pfnv{32.46}{2.20} & \pfnv{18.84}{1.70} \\
    DM-TabICLv2
      & \pfnv{1.05}{0.25}   & \pfnv{25.28}{1.30} & \pfnv{13.35}{0.81}
      & \pfnv{10.12}{0.76}   & \pfnv{44.05}{2.10} & \pfnv{30.10}{2.10} \\
    MIPS
      & \pfnv{0.27}{0.12}   & \pfnv{32.56}{35.00} & \pfnv{10.31}{0.76}
      & \pfnv{8.26}{4.00}   & \pfnv{27.45}{4.20} & \pfnv{18.86}{4.10} \\
    \midrule
    \multicolumn{7}{@{}l}{\textit{UltraFeedback}} \\
    \rowcolor{pfnshade}PFN-OPE (ours)
      & \pfnv{1.120}{0.041}  & \pfnb{5.01}{1.00} & \pfnb{9.86}{0.42}
      & \pfnb{0.038}{0.004} & \pfnb{6.22}{0.58} & \pfnb{6.52}{0.39} \\
    DM-ridge
      & \pfnv{1.55}{0.15}   & \pfnv{9.50}{0.92} & \pfnv{12.11}{0.56}
      & \pfnv{0.043}{0.043} & \pfnv{13.76}{1.40} & \pfnv{19.55}{1.10} \\
    DM-kernel
      & \pfnv{1.56}{0.15}   & \pfnv{8.71}{0.89} & \pfnv{12.24}{0.51}
      & \pfnv{0.046}{0.046} & \pfnv{14.42}{1.30} & \pfnv{20.18}{1.10} \\
    DM-MLP
      & \pfnv{1.59}{0.67}   & \pfnv{12.95}{2.20} & \pfnv{13.86}{2.70}
      & \pfnv{1.66}{1.20}   & \pfnv{18.88}{2.90} & \pfnv{24.08}{4.40} \\
    DM-TabPFN
      & \pfnv{2.49}{0.24}   & \pfnv{8.65}{0.79} & \pfnv{9.99}{0.68}
      & \pfnv{0.062}{0.040} & \pfnv{14.60}{1.60} & \pfnv{16.80}{1.10} \\
    DM-TabICLv2
      & \pfnb{0.230}{0.066}  & \pfnv{13.05}{1.10} & \pfnv{17.80}{0.69}
      & \pfnv{1.11}{0.23}   & \pfnv{20.07}{1.80} & \pfnv{30.16}{1.10} \\
    MIPS
      & \pfnv{1.31}{0.41}   & \pfnv{10.30}{1.50} & \pfnv{14.47}{0.48}
      & \pfnv{4.25}{3.80}   & \pfnv{15.80}{4.10} & \pfnv{24.62}{4.60} \\
    \bottomrule
  \end{tabular}
\end{table}

\subsection{PFN-OPE generalizes to four unseen rewards}
\label{sec:exp-reward}
\paragraph{Setup.}
Reward shift is the case in which the direct method must refit. We test whether PFN-OPE, which never saw the new reward, can replace that refit. The policies are the six ordered pairs of seen LLMs, and only the reward changes: the seen aggregate and the four unseen rewards, at $n=512$. The held-out ArmoRM rewards are new definitions from a reward model the prior knows. GRM is a reward model the prior never saw.
\vspace{-4pt}
\paragraph{Results.}
PFN-OPE has the lowest MSE under all five rewards on both datasets, by $2.0$ to $9.3$ times over the best baseline (\cref{fig:reward-shift}). The gain shrinks with distance from the prior but does not vanish. It is $3.7$ to $9.3$ times for the rewards built from ArmoRM's objectives, which the prior's reward combinations span, and $2.0$ to $4.1$ times under GRM, which lies outside them. The baselines see the new reward through the same logged rewards and refit to it. The gain is 3.7–9.3× for rewards spanned by the prior and 2.0–4.1× for out-of-prior GRM.

\begin{figure}[t]
\centering
\includegraphics[width=0.85\linewidth, trim=0 0 0 36pt, clip]{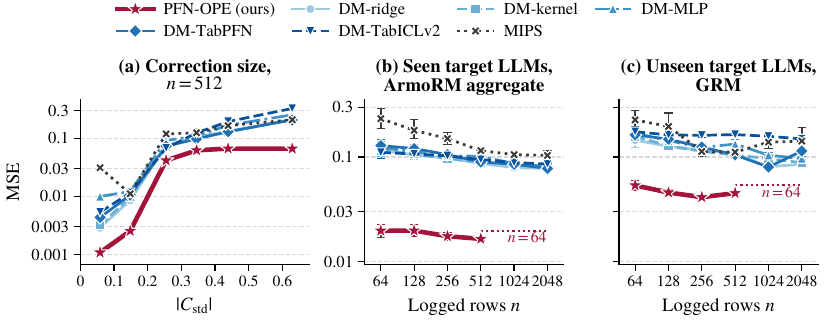}
\vspace{-4pt}
\caption{\textbf{The advantage grows with the correction and survives 32 times more logged data than the baselines.} Legend as in Figure~\ref{fig:reward-shift}. (a) MSE against the standardized correction $|C_{\mathrm{std}}|$ at $n=512$: above $0.2$, every baseline's MSE rises with the correction, PFN-OPE's stays lower. (b, c) MSE against logged rows $n$, for seen LLMs under ArmoRM (b) and unseen LLMs under GRM (c).}
\label{fig:gains}
\end{figure}

\subsection{PFN-OPE generalizes to unseen target LLMs under an unseen reward model}
\label{sec:exp-joint}

\paragraph{Setup.}
This is the deployment case of \cref{sec:intro}: a candidate LLM the prior never saw, scored by a reward model the prior never saw. The target policies are Llama-3.2-3B and Gemma-3-1B, and the reward is GRM, on both datasets: 12 behavior--target pairs at $n=512$.
\vspace{-4pt}
\paragraph{Results.}
PFN-OPE has the lowest MSE in 11 of 12 pairs (\cref{tab:joint-shift}). The gain is $3.1$ to $19$ times in five pairs on HelpSteer2 and $1.7$ to $2.6$ times in three pairs on UltraFeedback. In the other four pairs, the estimators are close. Two of these have an MSE below $0.001$ for PFN-OPE and the closest DM baselines: the logged mean is already near the target value, so there is little to correct. In the remaining two, both on UltraFeedback, PFN-OPE ties DM-TabPFN from Llama-3.2-1B to Llama-3.2-3B and loses to DM-TabICLv2 by $4.9$ times from Qwen2.5-7B to Llama-3.2-3B.

\begin{wrapfigure}[17]{r}{0.39\textwidth}
  \vspace{-\intextsep}
  \centering
  \includegraphics[width=\linewidth]{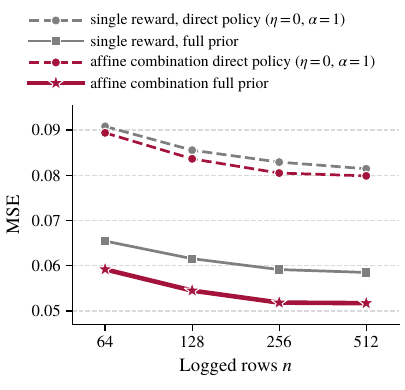}
  \caption{\textbf{Ablation of the prior.} MSE of four PFNs pretrained in a $2\times2$ design over the reward and the policy prior.}
  \label{fig:prior-ablation}
\end{wrapfigure}

\subsection{PFN-OPE is robust to large correction and less logged data}
\label{sec:exp-correction-n}

\paragraph{Setup.}
\Cref{fig:gains}a tests whether PFN-OPE performs in large correction $C$ settings. We pool all $n=512$ tasks of Q1 and Q2 and bin them by the correction in units of the logged reward's standard deviation, $|C_{\mathrm{std}}|$. For \Cref{fig:gains}b,c, we take the high-correction third of the logged datasets and grow $n$ from 64 to 2{,}048, for seen targets under the ArmoRM aggregate (b) and unseen targets under GRM (c). PFN-OPE is evaluated up to $n=512$.
\paragraph{Results.}
Below $|C_{\mathrm{std}}|=0.2$ every estimator is accurate, with MSE at or below $0.012$ for every direct method, because the logged mean is already close to the target value. Above $0.2$ the baselines' MSE grows with the correction and reaches $0.21$ to $0.32$ in the largest bin, while PFN-OPE's stays between $0.04$ and $0.07$ (\cref{fig:gains}a). The ratio to the best baseline widens from $1.6$ to $3.1$. A regressor fitted to the logged rewards predicts target rewards near them, so it inherits the logged mean's error $C^2$. An estimator trained to predict $C$ does not. More data does not close the gap. With 64 rows, PFN-OPE is more accurate than every baseline with 2{,}048, by $3.9$ times for seen targets and $1.6$ times for unseen targets (\cref{fig:gains}b,c). The baselines' curves flatten above PFN-OPE's, so their remaining error is due to extrapolation. In deployment, those rows are annotations: PFN-OPE with 64 rows is more accurate than the baselines with 32 times as many.

\subsection{Prior ablation: the policy axis of the prior carries the gain}
\label{sec:exp-ablation}

\vspace{-0.1cm}
\paragraph{Setup.}
The prior is the estimator's entire inductive bias (\cref{sec:problem}), and \cref{sec:prior} widened it along two axes: reward definitions and policy shifts. To see which axis carries the gain, we pretrained four PFNs in a $2\times2$ design: one reward function per task or the affine combination of \cref{eq:reward-prior}, crossed with the direct policies ($\eta=0$, $\alpha=1$) or the full policy prior with tilted and routed policies.
\vspace{-0.2cm}
\paragraph{Results.}
The full policy prior lowers MSE by 28\% to 36\% at every $n$, whether the reward prior is a single function or a combination (\cref{fig:prior-ablation}). Reward combinations help little on their own, improving by 2\% to 3\% with pool policies and by 10\% to 12\% on top of the full policy prior.

\subsection{Prior coverage is important in pretraining}
\label{sec:exp-coverage}

\begin{figure}[t]
  \centering
  \begin{minipage}[c]{0.75\linewidth}
    \centering
    \includegraphics[width=\linewidth]{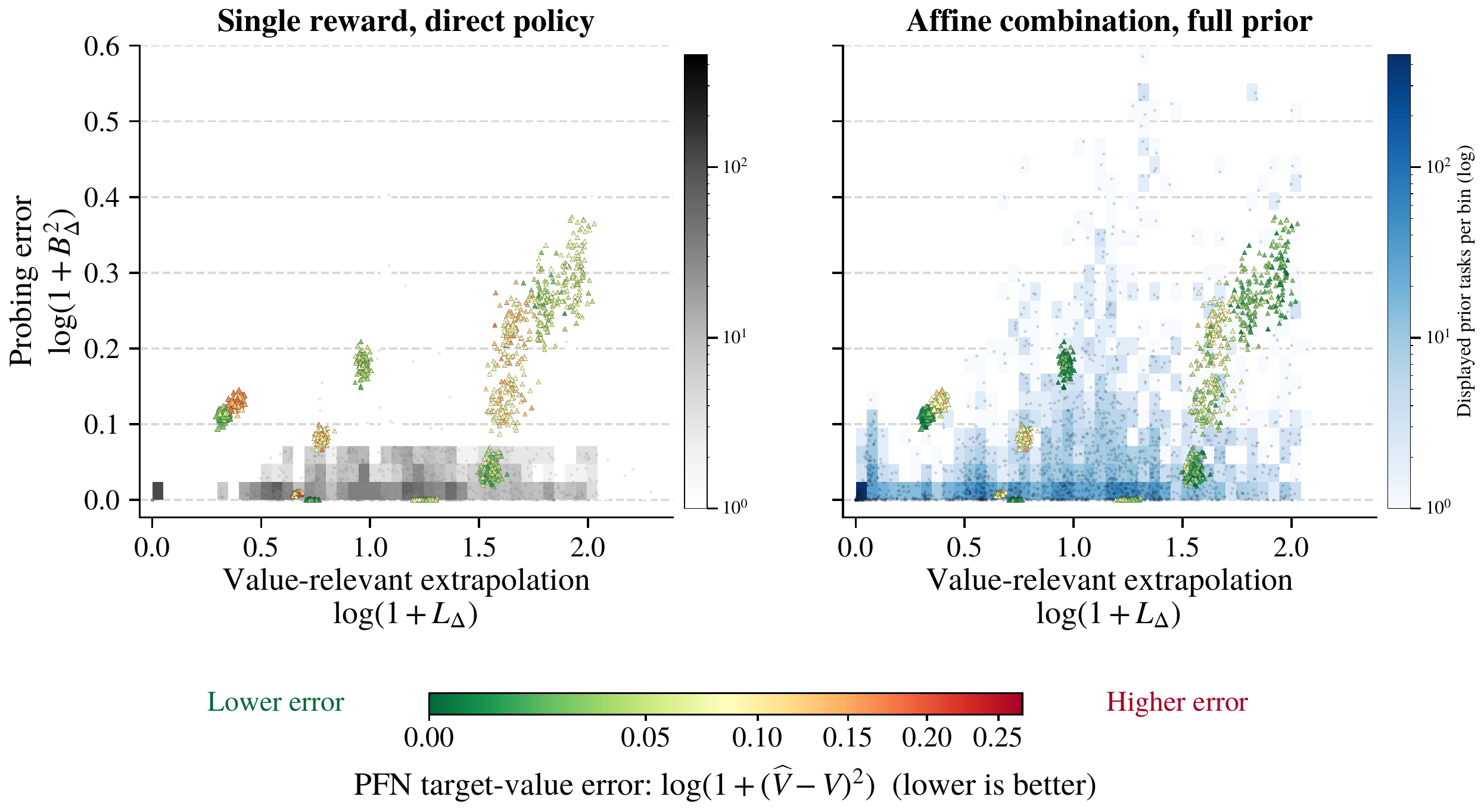}
  \end{minipage}\hfill
  \begin{minipage}[c]{0.23\linewidth}
    \caption{\textbf{The PFN fails where its prior has no tasks, and the shifted policies fill the gap.} Each panel shows the pretraining tasks. Triangles are the evaluation tasks with an unseen target LLM under GRM. Left: one reward per task, direct policies. Right: affine reward combinations, full policy prior.\vspace{-5pt}}
    \label{fig:prior-coverage}
  \end{minipage}
\end{figure}

\paragraph{Setup.}
We place every task in a plane whose two axes measure how hard the task is for a linear reward model on the frozen embeddings (definitions in \cref{app:coverage-diagnostics}). The horizontal axis, $L_\Delta$, is how far the target responses move away from the logged ones in embedding space, in directions the logged responses do not cover. The vertical axis, the probe correction error $B_\Delta^2$, is how wrong a linear reward model fitted to the logs is about the value gap between the two policies. We plot 1{,}000 pretraining tasks from each of the two arms of the ablation and the 1{,}000 evaluation tasks with an unseen target LLM under GRM, at $n=128$. Each evaluation task is colored by the error of the PFN pretrained on that panel's prior.
\vspace{-5pt}
\paragraph{Results.}
The two priors cover the same region except the left corner and upper region, and that corner is where the two PFNs differ (\cref{fig:prior-coverage}). The pool-only prior has no pretraining task there, and its PFN has its highest error on the evaluation tasks there. The full prior has tasks there, and its PFN's error there is low. Elsewhere, both PFNs are accurate where their priors are dense. Every evaluation task in the corner is a Llama-3.2-1B$\to$Llama-3.2-3B task: a target from the family of the logged LLM, whose responses differ in reward more than in form. This shift is closest to a post-trained checkpoint of the logged model, and the shifted policies are what put it in the prior.
\vspace{-4pt}

\vspace{-4pt}
\section{Conclusion}
\vspace{-4pt}
We introduced PFN-OPE, a prior-data fitted network that amortizes off-policy evaluation of LLMs. We pretrained it once on tasks built from a response pool, where every target value is known by construction. At test-time, it maps a logged dataset and sampled target responses to a value estimate in one forward pass, with no per-task fitting. On HelpSteer2 and UltraFeedback, PFN-OPE has $2.0$ to $9.3$ times lower MSE than the best baseline. It has the lowest MSE in 11 of 12 behavior--target pairs when the target LLM and the reward model are both unseen. As post-training objectives and feedback signals continue to shift, we see PFN-OPE as a step toward evaluation that keeps pace by reusing one pretraining run across the stream of target policies that deployment produces.

\subsection*{AI use statement}
In this work, we used generative AI tools to assist with the design of our
research methodology and experiments (providing feedback on experiments), to implement methods, to reformat datasets, and to support qualitative data analysis. Beyond the LLMs and reward models studied in this work, whose responses and scores form the response pool, we did not use generative AI tools to generate dataset develop theoretical models
or conceptual frameworks, propose or refine hypotheses, or interpret results,
and formulating or proving mathematical claims, writing proofs, and translation
are not applicable to this work. Additionally, we used generative AI tools to
suggest experimental parameters, create and edit software code, draft parts of
the paper, and edit the paper to improve readability. We have reviewed all
AI-assisted work. AI-generated code was reviewed and tested for correctness before being used in any experiment. AI-assisted qualitative analyses were independently verified by 2 authors. All AI-drafted or AI-edited text was checked by the authors for accuracy and revised as needed. We take responsibility for the final content of this work, including text, claims or artifacts produced with the aid of generative AI.

\bibliography{iclr2027_conference}
\bibliographystyle{iclr2027_conference}

\newpage
\appendix
\section*{Appendix Contents}
\addcontentsline{toc}{section}{Appendix Contents}
\startcontents[sections]
\printcontents[sections]{}{1}{\setcounter{tocdepth}{2}}
\clearpage
\section{Related Work}

\paragraph{Prior-fitted networks.}
Prior-fitted networks (PFN) train a transformer on datasets sampled from a synthetic prior so that a single forward pass approximates posterior-predictive inference in context \citep{muller2022transformers}. 
PFNs have been successfully applied for tabular prediction \citep{hollmann2023tabpfn,tabpfnv2,helli2024driftresilient}, Bayesian optimization \citep{muller2023pfns4bo} and causal effect estimation \citep{balazadeh2025causalpfn}. 
In our work, we extend the idea to off-policy evaluation of LLMs.

\paragraph{Off-policy evaluation from text embeddings.}
Classical OPE estimators such as IPS are unable to handle continuous or large action space. 
Kernel-based IPS methods relax target policy distributions in IPS computation to a kernel \citep{lee2024kernel,sachdeva2024off}. However, it is non-trivial to extend these kernel-based IPS to high dimensional text embedding.
MIPS \citep{saito2022off} applies IPS in a given embedding space, marginalizing over action variations that share the same embedding.
\citet{cief2023learning} proposed using reward prediction loss to train the embedding for MIPS.
However, these MIPS-based methods still need the per-dataset fit.

\paragraph{In-context bandits and reinforcement learning.}
In bandits and reinforcement learning settings, in-context learning has demonstrated efficient policy generalization \citep{xu2022prompting}. These methods condition transformer-based policy models on trajectory data or interaction history to generalize to new tasks. 
Our work applies this in-context paradigm to off-policy evaluation: we train PFN-OPE to perform in-context value estimation, eliminating per-task fitting while enabling generalization to unseen policies and reward functions. 
\section{Algorithms}
\label{app:algorithm}
\Cref{alg:pretraining} gives every step of the procedure summarized in \cref{alg:overview}.

\begin{algorithm}[t]
\caption{PFN-OPE pretraining}
\label{alg:overview}
\small
\begin{algorithmic}[1]
\Require Response pool: prompts $\mathcal X_0$, pool policies $\{\wh g\}_{g\in\mathcal G}$, stored rewards $R_1,\ldots,R_J$.
\For{each pretraining example}
    \State Sample a task $\tau$: prompt distribution $p(x\mid\tau)$, weights $\lambda_\tau,\omega_\tau$, LLMs $g,g'$, and $(\eta,\alpha)$; form $R_\tau$ (\cref{eq:reward-prior}), $\mu=\wh g$, and $\pi$ (\cref{eq:routing-policy}).
    \State Compute $V_\tau(\pi)$ over the pool (\cref{eq:exact-task-value}).
    \State Draw $n$ prompts $x_i\sim p(\cdot\mid\tau)$ and, for each, $a_i^\mu\sim\mu(\cdot\mid x_i)$, $a_i^\pi\sim\pi(\cdot\mid x_i)$, and $r_i^\mu=R_\tau(x_i,a_i^\mu)$; form $O=(\mathcal D_\mu,\mathcal D_\pi)$.
    \State Set $C=V_\tau(\pi)-\bar r$ (\cref{eq:correction}).
    \State Update $\theta$ on the loss of \cref{eq:training-loss} at $(O,C)$.
\EndFor
\end{algorithmic}
\end{algorithm}

\begin{algorithm}[h]
\caption{Generating pretraining data for PFN-OPE}
\label{alg:pretraining}
\small
\begin{algorithmic}[1]
\Require Prompts $\mathcal X_0$; LLMs $\mathcal G$; reward functions $R_1,\ldots,R_J$; responses per prompt $K$.
\Ensure Pretraining examples $(O,C)$, with the rewards of the target responses hidden from $O$.
\Statex \textbf{Stage 1: build the response pool once}
\For{each prompt $x\in\mathcal X_0$ and LLM $g\in\mathcal G$}
    \State Sample $K$ responses $a_g^{(1)}(x),\ldots,a_g^{(K)}(x)\sim g(\cdot\mid x)$ and store them as $\mathcal A_g(x)$.
    \State Compute $R_j(x,a)$ for every stored response $a\in\mathcal A_g(x)$ and every $j=1,\ldots,J$.
\EndFor
\State Standardize each reward over the pool, $\tilde R_j=(R_j-m_j)/s_j$.
\State Define the pool policy $\wh g(\cdot\mid x)$ of each LLM as the uniform distribution on $\mathcal A_g(x)$ (\cref{eq:pool-policy}).
\Statex \textbf{Stage 2: sample one task and one observation per pretraining example}
\For{each pretraining example}
    \State Draw a prompt distribution $p(\cdot\mid\tau)$ on $\mathcal X_0$ and the number of logged rows $n$.
    \State Draw evaluation weights $\lambda_\tau\sim\mathcal N\bigl(w,\operatorname{diag}(w)-ww^\top\bigr)$ and set $R_\tau(x,a)=\sum_{j=1}^{J}\lambda_{\tau j}\tilde R_j(x,a)$ (\cref{eq:reward-prior}).
    \State Draw selection weights $\omega_\tau\in\Delta^{J-1}$, independently of $\lambda_\tau$, and set $S_\tau(x,a)=\sum_{j=1}^{J}\omega_{\tau j}\tilde R_j(x,a)$.
    \State Draw a behavior LLM $g\in\mathcal G$ and a target LLM $g'\in\mathcal G$, and set $\mu=\wh g$.
    \State Draw the policy mechanism: pool policy ($\eta=0$, $\alpha=1$) with probability $0.4$, selected ($\eta>0$, $\alpha=1$) with probability $0.4$, routed ($\alpha<1$) with probability $0.2$; draw $\eta$ and $\alpha$ accordingly.
    \State Form the target policy (\cref{eq:routing-policy}),
    \Statex \hspace{\algorithmicindent}$\displaystyle
    \nu_\eta(a\mid x)=\frac{\wh g'(a\mid x)\,e^{\eta S_\tau(x,a)}}{\sum_{a'}\wh g'(a'\mid x)\,e^{\eta S_\tau(x,a')}},
    \qquad \pi=(1-\alpha)\,\mu+\alpha\,\nu_\eta$.
    \State Compute the target value over the pool (\cref{eq:exact-task-value}),
    \Statex \hspace{\algorithmicindent}$\displaystyle
    V_\tau(\pi)=\sum_{x\in\mathcal X_0}p(x\mid\tau)\sum_{a}\pi(a\mid x)\,R_\tau(x,a)$.
    \State Draw $n$ prompts $x_1,\ldots,x_n\sim p(\cdot\mid\tau)$.
    \For{$i=1,\ldots,n$}
        \State Draw $a_i^\mu\sim\mu(\cdot\mid x_i)$ and $a_i^\pi\sim\pi(\cdot\mid x_i)$ independently.
        \State Set the logged reward $r_i^\mu=R_\tau(x_i,a_i^\mu)$.
    \EndFor
    \State Form $\mathcal D_\mu=\{(x_i,a_i^\mu,r_i^\mu)\}_{i=1}^{n}$, $\mathcal D_\pi=\{(x_i,a_i^\pi)\}_{i=1}^{n}$, and $O=(\mathcal D_\mu,\mathcal D_\pi)$.
    \State Set $\bar r=\frac1n\sum_{i=1}^{n}r_i^\mu$ and $C=V_\tau(\pi)-\bar r$ (\cref{eq:correction}).
    \State Emit $(O,C)$; $O$ holds prompts, responses, and logged rewards only, and no task parameter.
\EndFor
\end{algorithmic}
\end{algorithm}

\section{Experimental details}
\label{app:experiments}
This appendix gives the evaluation protocol and the estimator settings behind \cref{sec:experiments}.

\subsection{Pool}
\label{app:pool}
\paragraph{Prompts and splits.}
Each dataset's prompts are split at random into a pretraining set and $1{,}024$ evaluation prompts. Every evaluation task in \cref{sec:experiments} draws its prompts from the evaluation set, so no evaluation prompt, behavior response, or target response appears in pretraining.

\paragraph{Response generation and scoring.}
For each prompt and each LLM, seen or unseen, we sample $K=8$ responses by nucleus sampling with temperature $0.7$ and top-$p$ $0.9$. Every response is scored against its prompt by ArmoRM and GRM.

\subsection{Evaluation protocol}
\label{app:protocol}
\paragraph{Correction and rank groups.}
For a task with logged rewards $r_1,\ldots,r_n$, the correction is the prediction target $C=V_\tau(\pi)-\bar r$ of \cref{eq:correction}. The standardized correction of \cref{sec:exp-correction-n} is $C_{\mathrm{std}}=C/s_\mu$, where $s_\mu$ is the population standard deviation of $r_1,\ldots,r_n$, floored at $10^{-6}$. $C$ is not the value gap $V_\tau(\pi)-V_\tau(\mu)$: it also contains the sampling deviation of $\bar r$, so it varies across tasks of one setting while $V_\tau(\pi)$ stays fixed. Within each setting (dataset, reward, behavior LLM, target LLM, and $n$), 100 candidate tasks were generated, ranked by $|C_{\mathrm{std}}|$.

\subsection{Estimators}
\label{app:estimators}
All estimators receive the same observation $O=(\mathcal D_\mu,\mathcal D_\pi)$; target rewards are used only to compute $V_\tau(\pi)$. Hyperparameters are fixed across all settings. DM-ridge, DM-kernel, and DM-MLP are refitted on every logged dataset. DM-TabPFN and DM-TabICLv2 use frozen pretrained weights. The TabPFN adapter projects the response embeddings $\phi_a$ of \cref{sec:architecture} to 128 dimensions with a fixed projection fitted on the response pool. MIPS estimates the density ratio between target and logged responses from their embeddings with prompt-grouped cross-fitting, and never sees a target reward. PFN-OPE's estimate is $\wh V_\theta(\pi;O)$ of \cref{eq:value-readout}: the logged mean plus the average of the 99 predicted quantiles of $C$.

\section{Prior-Coverage Diagnostics}
\label{app:coverage-diagnostics}
This appendix defines the two axes of \cref{fig:prior-coverage}. Both describe one fixed linear-ridge reward probe on the frozen embeddings of \cref{sec:architecture}; neither is a PFN-OPE prediction error.

\paragraph{Probe and features.}
For a task with logged data $\mathcal D_\mu$ of size $n=128$, the feature of a prompt--response pair is the concatenation of the unit-normalized prompt and response embeddings, $\phi(x,a)\defeq\tfrac{1}{\sqrt2}\,\bigl[\phi_x(x)/\|\phi_x(x)\|;\ \phi_a(a)/\|\phi_a(a)\|\bigr]$. A ridge regressor is fitted to the logged rewards in $F=4$ prompt-grouped folds, fold $b$ using the $96$ logged rows outside it, giving $\wh f_b(x,a)=\wh\beta_b^\top\phi(x,a)+\wh\gamma_b$; the probe is the fold average $\bar f\defeq\tfrac1F\sum_{b=1}^{F}\wh f_b$. The expectations $\E_\mu[\cdot]$ and $\E_\pi[\cdot]$ below are over the task's prompt distribution and its pool policies, computed exactly on the stored responses.

\paragraph{Probe correction error (vertical axis).}
The probe's estimate of the value gap is $\wh\Delta\defeq\E_\pi[\bar f]-\E_\mu[\bar f]$ and the true gap is $\Delta\defeq V_\tau(\pi)-V_\tau(\mu)$. The diagnostic is the squared error of the estimate in units of the logged reward variance $\sigma_\mu^2\defeq\Var_\mu(R_\tau)$,
\begin{equation}
B_\Delta^2\defeq\frac{(\wh\Delta-\Delta)^2}{\sigma_\mu^2}.
\label{eq:probe-correction-error}
\end{equation}
Writing $e\defeq\bar f-R_\tau$ for the probe's pointwise error, $B_\Delta^2=(\E_\pi[e]-\E_\mu[e])^2/\sigma_\mu^2$: an error common to logged and target responses cancels, so $B_\Delta^2$ is zero when the probe misjudges both values by the same amount and large only when its error differs between the two response sets. It mixes finite-sample error, regularization, and misspecification of the linear probe, and it is not the probe's error on the target value itself.

\paragraph{Value-relevant extrapolation (horizontal axis).}
Let $d\defeq\E_\pi[\phi]-\E_\mu[\phi]$ be the mean feature shift and $\wh\Sigma_b$ the covariance of the features of fold $b$'s $96$ logged rows. The diagnostic is the ridge-regularized leverage of the shift, averaged over folds,
\begin{equation}
L_\Delta\defeq\frac1F\sum_{b=1}^{F} d^\top\bigl(\wh\Sigma_b+\lambda_b I\bigr)^{-1}d,
\label{eq:value-relevant-extrapolation}
\end{equation}
with $\lambda_b$ the ridge penalty of fold $b$. In the eigenbasis $\{(s_k,u_k)\}$ of $\wh\Sigma_b$, $d^\top(\wh\Sigma_b+\lambda_b I)^{-1}d=\sum_k (u_k^\top d)^2/(s_k+\lambda_b)$, so a shift along a direction in which the logged features barely vary (small $s_k$) counts more than the same shift along a direction they cover. The shift $d$ is what a linear probe's value correction depends on, $\wh\Delta=\bigl(\tfrac1F\sum_b\wh\beta_b\bigr)^\top d$, which is why we call the leverage value-relevant.

\section{Limitations}
\label{sec:limitations}
\paragraph{Evaluation targets are other LLMs.}
Every evaluation target is the pool policy of one LLM, so the policy shifts we test are changes of LLM. The prior-coverage analysis (\cref{sec:exp-coverage}) shows that the tilted policies reach the kind of task a post-trained checkpoint would pose.

\paragraph{Rewards come from reward models.}
Every reward in pretraining and evaluation is the output of ArmoRM or GRM, so the setting is semi-synthetic: reward definitions are heads and combinations of reward models rather than human rubrics, and any bias of those models enters the pool. Human-rated logs would test the reward shift the introduction describes; we did not have them at the scale the pool requires.

\paragraph{The prior is built from three LLMs.}
Transfer to an unseen checkpoint of a seen family and to one unseen family (\cref{sec:exp-joint}) is the evidence that the estimator generalizes beyond the pool's LLMs, and one unseen family is limited evidence. Widening the pool to more families and to actual post-training checkpoints is the natural extension, at the one-time cost of generating and scoring their responses.

\end{document}